\documentclass{llncs}
\usepackage{times}
\usepackage{graphicx}
\usepackage{amsmath}
\usepackage[usenames,dvipsnames]{xcolor}
\DeclareGraphicsExtensions{.jpg, .pdf, .png}
\usepackage{hyperref}
\usepackage{url}

\usepackage{mathtools}
\usepackage{bm}
\usepackage{esvect}

\usepackage{graphicx}
\usepackage{amsmath}

\DeclareGraphicsExtensions{.jpg, .pdf, .png}

\title{Partial Knowledge in Embeddings}

\author{
R.V.Guha}
\institute{  Schema.org \\
  \texttt{guha@guha.com}\\
}

\begin{document}

\maketitle

\begin{abstract}

Representing domain knowledge is crucial for  any task. There has been
a wide range of techniques developed to represent this knowledge,  from older
logic based approaches to the more recent deep learning based techniques
(i.e. embeddings). In this paper, we discuss some of these
methods, focusing on the representational expressiveness tradeoffs
that are often made. In particular,  we focus on the the ability of
various techniques to encode `partial knowledge' - a key component of
successful knowledge systems. We introduce
and describe the concepts of \textit{ensembles of
  embeddings} and \textit{aggregate embeddings} 
and demonstrate how they allow for partial knowledge.

\end{abstract}

 \section*{Motivation}

 Knowledge about the domain is essential to performing any task.
Representations of this knowledge have ranged over a broad
 spectrum in terms of the features and tradeoffs. Recently, with the
 increased interest in deep neural networks, work has focussed on
 developing knowledge representations based on the kind of structures
used in such networks. In this paper, we discuss some of the
representational expressiveness tradeoffs that are made, often
implicitly. In particular we focus on the loss of the ability to
encode partial knowledge and explore two different paths to regain
this ability.

\subsection*{Logic Based Representations}

 Beginning with McCarthy's Advice Taker \cite{advicetaker}, logic has
 been the formal foundation for a wide range of knowledge
 representations. These have ranged across the spectrum of formal
 models to the more experimental languages created as part of working
 systems. The more formal work started with McCarthy and Hayes's 
  Situation Calculus \cite{sitcalc}, the various flavors of
 non-monotonic logics \cite{reiter}, \cite{circumscription} and
 included proposed
 axiomatizations such as those suggested in the `Naive physics
 manifesto' \cite{naivephysics} and Allen's temporal representation
 \cite{allen}. There were a number of less formal approaches, that
 also had their roots in logic. Notable amongst these include Minsky's
 frames \cite{frames}, EMycin \cite{emycin}, KRL \cite{krl} and
 others. The representation language used by Cyc, CycL \cite{cyc} is a
 hybrid of these approaches.

Most of these representation systems can be formalized as variants of
first order logic. Inference is done via some form of theorem
proving. One of the main design issues in these systems is the
tradeoff between expressiveness and inferential complexity. 

More recently, systems such as Google's `Knowledge Graph' have started
finding use, though their extremely limited expressiveness makes them
much more like simple databases than knowledge bases.

Though a survey of the wide range of KR systems that have been built
is beyond the scope of this paper, we note that there are some very
basic abilities all of these systems have. In particular, 
\begin{itemize}
\item They are all relational, i.e., are built around the concept of
  entities that have relations between them
\item They can be updated incrementally
\item The language for representing 'ground facts' is the same as the
  language for representing generalizations. There is no clear line
  demarcating the two.
\item They can represent partial knowledge. It is this feature we
  focus on in this paper. 
\end{itemize}

\subsection*{Feature Vectors}

One of the shortcomings of the logic based representations was
that they largely assumed that everything the system knew was
manually given to the system, i.e., learning from ground data was not
a central part of the design. The complexity of the representational
formalisms and wide range of possible functions or expressions have made
machine learning rather difficult in traditional knowledge
representation systems.

The rise of machine learning as the primary mechanism for creating
models of the domain have lead to the use of much simpler
representations. In particular, we notice,

\begin{itemize}
\item The language for representing ground facts is distinct from the
  language for representing generalizations or models. The ground
  facts about the domain, i.e., the training data, is usually
  represented as a set of feature vectors.

\item The language for representing generalizations or models is also
  usually highly restricted. Each family of models (e.g., linear
  regression, logistic regression, support vector machines, has a 
function template with a number of parameters that the learning
algorithm computes. Recent work on neural networks attempts to capture
the generality of Turing machines with deep networks, but the
structure of the function learnt by these systems is still uniform.

\item The language for representing ground facts is propositional,
  i.e., doesn't have the concept of entities or relations. This
  constraint makes it very difficult to use these systems for modeling
  situations that are relational. Many problems, especially those that
  involve reasoning about people, places, events, etc. need the
  ability to represent these entities and the relations between them.

\item Most of these systems allow for partial knowledge in their
  representation of ground facts. i.e., some of the features in the
  training data may be missing for some of the instances of the
  training data.

\end{itemize}
The inability of feature vectors to represent entities and relations
between them has lead to work in embeddings, which try to represent
entities and relations in a language that is more friendly to learning
systems. However, as we note below, these embedding based
representations leave out an important feature of classical logic
based representations --- a feature we argue is very important.

We first review embedding based representations, show how they are
incapable of capturing partial information

\subsection*{Embeddings}

Recent work on distributed representations [\cite{socher}, \cite{manning}, 
\cite{bordes}, \cite{bordes2014semantic}, \cite{quoc}] has explored the use of embeddings as a representation
tool. These approaches typically 'learn an embedding', which maps terms and statements
in a knowledge base (such as Freebase \cite{freebase}) to points in an N-dimensional vector space.
Vectors between points can then be interpreted as relations between
the terms. A very attractive property of these distributed
representations is the fact that they are learnt from a set of
examples.

Weston, Bordes, et. al. \cite{bordes} proposed a simple model (TransE)
wherein each entity is mapped to a point in the N-dimensional space
and each relation is mapped to a vector. So, given the triple $r(a,
b)$, we have the algebraic constraint $\overrightarrow{\boldsymbol{a}}
- \overrightarrow{\boldsymbol{b}} =
\overrightarrow{\boldsymbol{r}} + \epsilon$, where $\epsilon$ is an error term. 
Given a set of ground facts, TransE picks coordinates for
each entity and vectors for each relation so as to minimize the
cumulative error. This simple formulation has some 
problems (e.g., it cannot represent many to many relations), which has
been fixed by subsequent work (\cite{wang2014knowledge}). However, the
core representational deficiency of TransE has been retained by these
subsequent systems.

The goal of systems such as TransE is to learn an embedding that 
can predict new ground facts from old ones. They do this by
dimensionality reduction, i.e., by using a low number of dimensions
into which the the ground facts are embedded. Each triple is mapped 
into an algebraic constraint of the form $\overrightarrow{\boldsymbol{a}} -\overrightarrow{\boldsymbol{b}} = \overrightarrow{\boldsymbol{r}} + \epsilon$ and an
optimization algorithm is used determine the vectors for the objects
and relations such that the er- ror is minimized. If the number 
of dimensions is sufficiently large, the $\epsilon$s are all zero and 
no generalizations are made. As the number of dimensions is reduced, 
the objects and relation vectors get values that minimize the εs, in
effect learning generalizations. Some of the generalizations learnt 
may be wrong, which contribute to the non-zero $\epsilon$s. As the number of 
dimensions is reduced further, the number of wrong generalizations
increases. This is often referred to as `KB completion'.

We believe that the value of such embeddings goes beyond 
learning simple generalizations. These embeddings are a representation
of the domain and should be usable by an agent to encode its knowledge
of the domain. Further, any learning task that takes
descriptions of situations that are best represented using a graph now
has a uniform representation in terms of this embedding. When it is
used for this purpose, it is very important that the embedding
accurately capture what is in the training data (i.e., the input graph).
In such a case, we are willing to forgo learning in order to minimize
the overall error and can pick the smallest number of dimensions that
accomplishes this. In the rest of this paper, we will focus on this case.

\subsection*{Ignorance or Partial Knowledge}
In a logic based system that is capable of representing some
proposition $P$ (relational or propositional), it is trivial for the
system to not know whether $P$ is true or not. I.e., its knowledge
about the world is partial with respect to $P$.

However, when a set of triples is converted to an embedding, this
ability is lost. Consider a KB with the entities $Joe$, $Bob$,
$Alice$, $John$, $Mary$. It has a single
relation $friend$. The KB specifies that $Joe$ and $Bob$ are friends
and that $Alice$ and $John$ are friends and that $Mary$ and $John$ are
$not$ friends. It does not say anything about whether $Mary$ and
$Alice$ are friends or not friends. This KB can be said to have
partial knowledge about the relation between $Mary$ and $Alice$. When
this KB is converted into an embedding, to represent the agent's
knowledge about the domain, it is important that this aspect of the KB be
preserved.

Unfortunately, in an embedding, $Mary$ and $Jane$ are assigned
particular coordinates. Either $\overrightarrow{\boldsymbol{Mary}} - \overrightarrow{\boldsymbol{Jane}}$ is equal to
$\overrightarrow{\boldsymbol{friend}}$ or it is not. If it is, then, according to the
embedding, they are friends and if it is not, then, according to the
embedding, they are not friends. The embedding is a complete world,
i.e., every proposition is either true or false. There is no way of
encoding 'unknown' in an embedding. 

If the task is knowledge base completion, especially of an almost
complete knowledge base, then it may be argued that this deficiency is
excusable. However, if the KB is very incomplete (as most real world
KBs are) and if such as KB is being used as input training data, or as
the basis for an agents engagement with the world, this
forced completion could be problematic.

We now explore two alternatives for encoding partial knowledge in embeddings.

\subsection*{Encoding Partial Knowledge}

 Logic based formalisms distinguish between a knowledge base (a set of
 statements) and what it might denote. The object of the denotation is
 some kind of structure (set of entities and ntuples in the case of
 first order logic or truth assignments in the case of propositional
 logic). The KB corresponds not to a single denotation, but set of
 $possible$ denotations. Something is true in the KB if it holds in
 every possible denotation and false if it does not hold in any of the
 possible denotations. If it holds in some denotations and does not
 hold in some, then it is neither true nor false in the KB.

\subsubsection*{Ensemble of Embeddings}
In other words, the key in logic based systems to partial knowledge 
is the distinction between the KB and its denotation and the use 
of a set of possible denotations of a KB. 

Note that in logic based KR systems, these possible denotations (or possible
worlds) are almost always in the meta-theory of the system. They
are rarely actually instantiated. We could also imagine a KR system
which does instantiate an ensemble (presumbaly representative) 
of possible denotations and determine if a proposition is true, false
or unknown based on whether it holds in all, none of some of these
models. We  follow this approach with embeddings to encode partial knowledge.
Instead of a single embedding, we can use an
ensemble of embeddings to capture a KB. If we use a sufficient number
of dimensions, we should be able to create embeddings that have zero
cumulative error. Further, different initial conditions for the
network will give us different embeddings. These different embeddings
correspond to the different possible denotations. A ensemble of such
embeddings can be used to capture partial knowledge, to the extent
desired.

While this approach is technically correct in some sense, it also
defeats the purpose. The reason for creating the embeddings was in
part to create an encoding that could be used as input to a learning
system. When we go from a single embedding to an ensemble of
embeddings, the learning algorithm gets substantially complicated. 

One approach to  solving this problem is to develop a more compact
encoding for an ensemble of embeddings.  Remember that we don't need
to capture every single possible embedding corresponding to the given
KB. All we need is sample that is enough to capture the essential
aspects of the partiality of the KB.

\subsubsection*{Aggregate Models} Consider the set of points across different
models corresponding to a particular term. Consider a cluster
of these points (from an ensemble of embeddings)
which are sufficiently close to each other. This cluster or cloud 
of points (each of which is in a different embeddingl), corresponds to $an$ aggregate
of possible interpretations of the term. We can extend this approach for 
all the terms in the language. We
pick a subset of models where every term forms such a cluster. The set of clusters
and the vectors between  them gives us the aggregate model. Note that
in vectors corresponding to relations will also allow amount of variation.
If a model satisfies the KB, any linear transform of the model will
also satisfy the KB. In order to keep these transforms from taking
over, no two models that form an aggregate should be linear transforms
of each other.

In both aggregate  models, each object corresponds to a cloud in the 
N-dimensional space and the relation between objects is captured by their approximate
relative positions. The size of the cloud corresponds to the 
vagueness/approximateness (i.e., range of possible meanings) of the
concept.

 Partial knowledge is captured by the fact that while some of
the points in the clouds corresponding to a pair of terms may have
coordinates that maps to a given relation, other points in the clouds
might not. In the earlier example, we now have a set of points
corresponding to $Mary$ and $Jane$. Some of these are such that
$\overrightarrow{\boldsymbol{Mary}} - \overrightarrow{\boldsymbol{Jane}} = \overrightarrow{\boldsymbol{friend}}$, while others are such that
$\overrightarrow{\boldsymbol{Mary}} - \overrightarrow{\boldsymbol{Jane}} \ne \overrightarrow{\boldsymbol{friend}}$. Thus, these aggregates can
encode partial knowledge.

\subsection*{Conclusions}

  The ability to encode partial knowledge is a very important aspect
  of knowledge representation systems. While recent advances to KR
  using embeddings offer many attractions, the current approaches are
  lacking in this important aspect. We argue that an agent should be
  aware of what it doesn't know and should use representations that
  are capable of capturing this. We described one possible approach to
  extending embeddings to capture partial knowledge. While there is
  much work to be done before embeddings can be used as practical
  knowledge representation systems, we believe that with additions
  like the one described here, embeddings may turn out to be another
  useful addition to the knowledge representation tool chest.

\bibliographystyle{unsrt}
\bibliography{emt}

\begin{thebibliography}{10}

\bibitem{advicetaker}
John McCarthy.
\newblock {\em Programs with common sense}.
\newblock RLE and MIT Computation Center, 1960.

\bibitem{sitcalc}
John McCarthy and Patrick~J Hayes.
\newblock Some philosophical problems from the standpoint of artificial
  intelligence.
\newblock {\em Readings in artificial intelligence}, pages 431--450, 1969.

\bibitem{reiter}
Raymond Reiter.
\newblock Nonmonotonic reasoning.
\newblock {\em Annual review of computer science}, 2(1):147--186, 1987.

\bibitem{circumscription}
John McCarthy.
\newblock Circumscription—a form of non-monotonic reasoning.
\newblock {\em Artificial intelligence}, 13(1):27--39, 1980.

\bibitem{naivephysics}
Patrick~J Hayes et~al.
\newblock The naive physics manifesto.
\newblock 1978.

\bibitem{allen}
James~F Allen.
\newblock Maintaining knowledge about temporal intervals.
\newblock {\em Communications of the ACM}, 26(11):832--843, 1983.

\bibitem{frames}
Marvin Minsky.
\newblock Frames.
\newblock {\em the society of mind}, 1988.

\bibitem{emycin}
William van Melle, Edward~H Shortliffe, and Bruce~G Buchanan.
\newblock Emycin: A knowledge engineer’s tool for constructing rule-based
  expert systems.
\newblock {\em Rule-based expert systems: The MYCIN experiments of the Stanford
  Heuristic Programming Project}, pages 302--313, 1984.

\bibitem{krl}
Daniel~G Bobrow and Terry Winograd.
\newblock An overview of krl, a knowledge representation language.
\newblock {\em Cognitive science}, 1(1):3--46, 1977.

\bibitem{cyc}
Douglas~B Lenat, Ramanathan~V. Guha, Karen Pittman, Dexter Pratt, and Mary
  Shepherd.
\newblock Cyc: toward programs with common sense.
\newblock {\em CACM}, 33(8):30--49, 1990.

\bibitem{socher}
Richard Socher, Brody Huval, Christopher~D Manning, and Andrew~Y Ng.
\newblock Semantic compositionality through recursive matrix-vector spaces.
\newblock In {\em Proceedings of the 2012 Joint Conference on Empirical Methods
  in Natural Language Processing and Computational Natural Language Learning},
  pages 1201--1211. Association for Computational Linguistics, 2012.

\bibitem{manning}
Samuel~R Bowman, Christopher Potts, and Christopher~D Manning.
\newblock Recursive neural networks for learning logical semantics.
\newblock {\em arXiv preprint arXiv:1406.1827}, 2014.

\bibitem{bordes}
Antoine Bordes, Jason Weston, Ronan Collobert, Yoshua Bengio, et~al.
\newblock Learning structured embeddings of knowledge bases.
\newblock In {\em AAAI}, 2011.

\bibitem{bordes2014semantic}
Antoine Bordes, Xavier Glorot, Jason Weston, and Yoshua Bengio.
\newblock A semantic matching energy function for learning with
  multi-relational data.
\newblock {\em Machine Learning}, 94(2):233--259, 2014.

\bibitem{quoc}
Quoc~V Le and Tomas Mikolov.
\newblock Distributed representations of sentences and documents.
\newblock {\em arXiv preprint arXiv:1405.4053}, 2014.

\bibitem{freebase}
Kurt Bollacker, Colin Evans, Praveen Paritosh, Tim Sturge, and Jamie Taylor.
\newblock Freebase: a collaboratively created graph database for structuring
  human knowledge.
\newblock In {\em 2008 ACM SIGMOD}, pages 1247--1250. ACM, 2008.

\bibitem{wang2014knowledge}
Zhen Wang, Jianwen Zhang, Jianlin Feng, and Zheng Chen.
\newblock Knowledge graph embedding by translating on hyperplanes.
\newblock In {\em Proceedings of the Twenty-Eighth AAAI Conference on
  Artificial Intelligence}, pages 1112--1119, 2014.

\end{thebibliography}

\end{document}